\documentclass[runningheads]{llncs}
\usepackage[T1]{fontenc}
\usepackage{graphicx}
\usepackage{hyperref}
\usepackage{color}

\usepackage{todonotes}

\newcounter{xaviercounter}

\newcounter{clementcounter}

\newcounter{corentincounter}

\usepackage{csquotes}

\begin{document}
\renewcommand{\floatpagefraction}{.8}%
\title{Evolving Reservoirs \\ for Meta Reinforcement Learning}
%
%

\author{%
Corentin Léger\textsuperscript{*}\textsuperscript{\dag} \inst{1,2}  \and 
Gautier Hamon\textsuperscript{*}\inst{1} \and 
Eleni Nisioti\inst{1} \and 
Xavier Hinaut\textsuperscript{\ddag} \inst{2} \and 
Clément Moulin-Frier\textsuperscript{\ddag} \inst{1} 
}%
\institute{
Flowers Team, Inria and Ensta ParisTech, France.
\and
Mnemosyne Team, Inria; LaBRI, Univ. Bordeaux, Bordeaux INP, CNRS UMR 5800; Univ. Bordeaux, CNRS, IMN, UMR 5293, Bordeaux, France}




\authorrunning{C. Leger et al.}


\maketitle              

\def\thefootnote{*}\footnotetext{Equal first authors}
\def\thefootnote{‡}\footnotetext{Equal last authors}
\def\thefootnote{†}\footnotetext{Work done as intern at Flowers and Mnemosyne}

\begin{abstract}

Animals often demonstrate a remarkable ability to adapt to their environments during their lifetime. They do so partly due to the evolution of morphological and neural structures. These structures capture features of environments shared between generations to bias and speed up lifetime learning. In this work, we propose a computational model for studying a mechanism that can enable such a process. We adopt a computational framework based on meta reinforcement learning as a model of the interplay between evolution and development. At the evolutionary scale, we evolve reservoirs, a family of recurrent neural networks that differ from conventional networks in that one optimizes not the synaptic weights, but hyperparameters controlling macro-level properties of the resulting network architecture.  
At the developmental scale, we employ these evolved reservoirs to facilitate the learning of a behavioral policy through Reinforcement Learning (RL). Within an RL agent, a reservoir encodes the environment state before providing it to an action policy.
We evaluate our approach on several 2D and 3D simulated environments. Our results show that the evolution of  reservoirs can improve the learning of diverse challenging tasks. We study in particular three hypotheses: the use of an architecture combining reservoirs and reinforcement learning could enable (1) solving tasks with partial observability, (2) generating oscillatory dynamics that facilitate the learning of locomotion tasks, and (3) facilitating the generalization of learned behaviors to new tasks unknown during the evolution phase.

\end{abstract}

\section{Introduction} \label{intro}

Animals demonstrate remarkable adaptability to their environments, a trait honed through the evolution of their morphological and neural structures \cite{tierney1995evolutionary,pearson2000neural}. They are born equipped with both hard-wired behavioral routines (e.g. breathing, motor babbling) and learning capabilities for adapting based on their own experience. The costs and benefits of evolving hard-wired behaviors vs. learning capabilities depend on different factors, a central one being the level of unpredictability of environmental conditions across generations \cite{stephens1991ChangeRegularityValue,johnstonSelectiveCostsBenefits1982}. Environmental challenges that are shared across many generations favor the evolution of hard-wired behavior (e.g. breathing). On the other hand, traits whose utility can hardly be predicted from its utility in previous generations are likely to be learned through individual development (e.g. learning a specific language). Some brain regions might have evolved to generically facilitate the learning of diverse behaviors. For example, central pattern generators (CPGs) enable limb bambling, which may facilitate locomotion, pointing and vocalizations in humans \cite{marder2001central}. Another example is the prefrontal cortex (PFC), a brain region that maps inputs within a high-dimensional non-linear space from which they can be decoded by other brain regions, acting as a reservoir for computations \cite{mante2013context,hinaut2011}.  

\begin{figure}
\begin{center}
\includegraphics[width=0.9\textwidth]{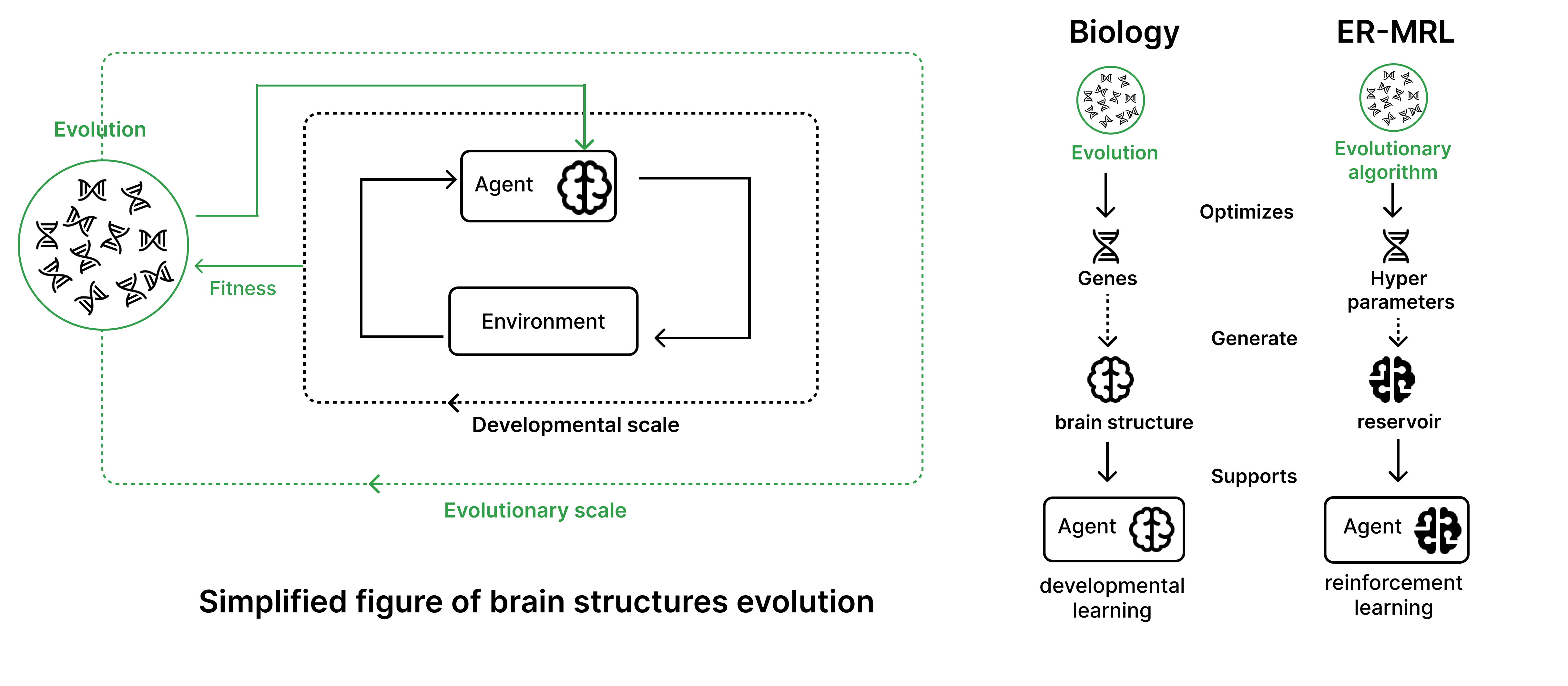}
\caption{(left) A simplified view of the evolution of brain structures. The generating parameters of neural structures are modified at an evolutionary loop. In the developmental loop, agents equipped with these neural structures learn to interact with their environment (right) Parallel to our computational approach.  We propose a computational framework where an evolutionary algorithm optimizes hyperparameters that generate neural structures called reservoirs. These reservoirs are then integrated into RL agents that learn an action policy to maximize their reward in an environment} \label{intro_fig}
\end{center}

\end{figure}

This prompts an intriguing question: How can neural structures, optimized at an evolutionary scale, enhance the capabilities of agents to learn complex tasks at a developmental scale? To address this question, we propose to model the interplay between evolution and development as two nested adaptive loops: neural structures are optimized through natural selection over generations (i.e. at an evolutionary scale), while learning specific behaviors occurs during an agent’s lifetime (i.e. at a developmental scale). Fig.~\ref{intro_fig} illustrates the interactions between evolutionary-scale and developmental-scale optimization. 
This model agrees with recent views on evolution that emphasize the importance of both scales for evolving complex skills \cite{laland_extended_2015,kauffman_origins_1993}.
It is also compatible with the biological principle of a \emph{genomic bottleneck}, i.e. the fact that the information contained in the genome of most organisms is not sufficient to fully describe their morphology~\cite{zador2019critique}.
In consequence, genomes must instead encode macro-level properties of morphological features such as synaptic connection patterns.

In line with these biological principles, we propose a novel computational approach, called Evolving Reservoirs for Meta Reinforcement Learning (ER-MRL), integrating mechanisms from Reservoir Computing (RC), Meta Reinforcement Learning (Meta-RL) and Evolutionary Algorithms (EAs).
We use RL as a model of learning at a developmental scale \cite{nussenbaum2019reinforcement,doya2007reinforcement}.
In RL, an agent interacts with a simulated environment through actions and observations, receiving rewards according to the task at hand.
The objective is to learn an action policy from experience, mapping the observations perceived by the agent to actions in order to maximize cumulative reward over time.
The policy is usually modeled as a deep neural network which is iteratively optimized through gradient descent.
We use RC as a model of how a genome can encode macro properties of the agent's neural structure.
In RC, the connection weights of a recurrent neural network (RNN) are generated from a handful of hyperparameters (HPs) controlling macro-level properties of the network related to connectivity, memory and sensitivity.
Our choice of using RC relies on its parallels with biological brain structures such as CPGs and the PFC \cite{hinaut2013,wyffels2009design}, as well as on the fact that its indirect encoding of a neural network in global hyperparameters makes it compatible with the genomic bottleneck principle mentioned above. 
Being a cheap and versatile computational paradigm, RCs may have been favored by evolution \cite{seoane2019evolutionary}.

We use Meta-RL to model how evolution shapes development \cite{clune2019ai,pedersen2021evolving}. Meta-RL considers an outer loop, akin to evolution, optimizing HPs of an inner loop, akin to development. At the evolutionary scale, we use an evolutionary algorithm to optimize a genome specifying HPs of reservoirs.
At a developmental scale, 
an agent equipped with a generated reservoir learns an action policy to maximize cumulative reward in a simulated environment.
Thus, the objective of the outer evolutionary loop is to optimize hyperparameters of reservoirs in order to facilitate the learning of an action policy in the inner developmental loop.

Using this computational model, we run experiments in diverse simulated environments, e.g. 2D environments where the agent learns how to balance a pendulum and 3D environments where the agent learns how to control complex morphologies.
These experiments provide support to three main hypotheses for how evolved reservoirs can affect development.
First, they can facilitate solving partially-observable tasks, where the agent lacks access to all the information necessary to solve the task.
In this case, we test the hypothesis that the recurrent nature of the reservoir will enable inferring the unobservable information. Second, it can generate oscillatory dynamics useful for solving locomotion tasks.
In this case, the reservoir acts as a meta-learned CPG.
Third, it can facilitate the generalization of learned behaviors to new tasks unknown during the evolution phase, a core hypothesis in meta-learning.
In our case, our expectation is that HPs of reservoirs evolved across different environments will capture some abstract properties useful for adaptation.

In Section~\ref{methods}, we detail the methods underlying our proposed model, including RL (Section~\ref{RL}), Meta-RL (Section~\ref{meta-RL}), RC (Section~\ref{RC}) and EAs (Section~\ref{EAs}). We then explain their integration into our ER-MRL architecture (Section~\ref{our_method}). Our results, aligned with the three hypotheses, are presented in Section~\ref{results}. Computational specifics and supplementary experiments can be found in the appendix. The source code and videos are accessible at \href{https://github.com/corentinlger/ER-MRL}{this link}.

\section{Background} \label{methods}

\begin{figure}
\begin{center}
\includegraphics[width=0.9\textwidth]{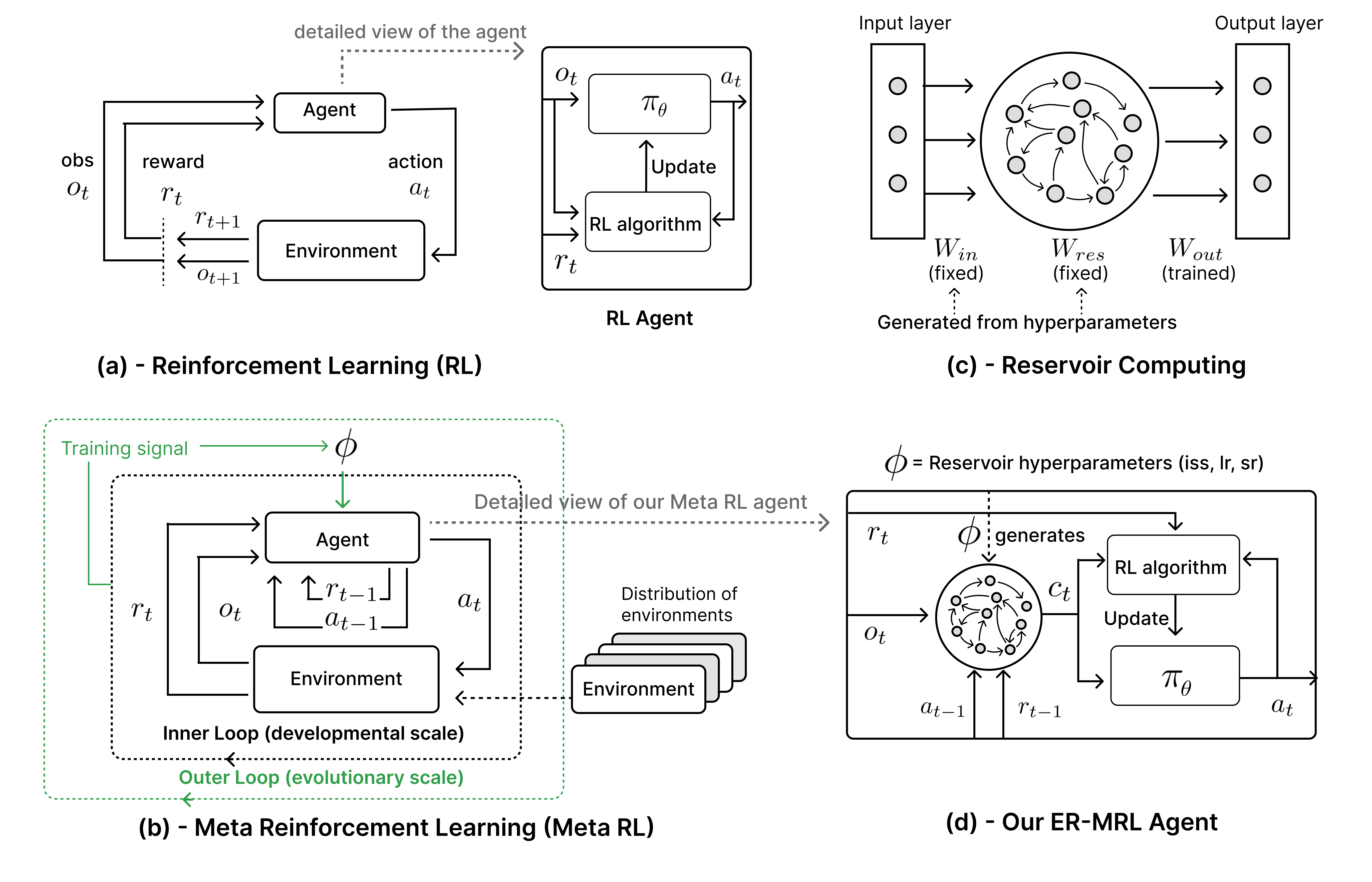}
\caption{Our proposed architecture, called ER-MRL, integrates several ML paradigms. We consider an RL agent learning an action policy (a), having access to a reservoir (c). We consider two nested adaptive loops in the spirit of Meta-RL (b). Our proposed architecture (d) consists in evolving HPs $\phi$ for the generation of reservoirs in an outer loop. In an inner loop, the agent learns an action policy, that takes as input the neural activation of the reservoir. The policy is trained using RL in order to maximize episodic return. Section~\ref{methods} provides the computational details of each ML paradigm.} \label{methods_fig}
\end{center}
\end{figure}


 \subsection{Reinforcement Learning as a model of development} \label{RL}

 Reinforcement Learning (RL) involves an agent that interacts with an environment by taking actions, receiving rewards, and learning an action policy in order to maximize its accumulated rewards (Fig.~\ref{methods_fig}.a).
 This interaction is formalized as a Markov Decision Process (MDP) \cite{puterman1990markov}.
 An MDP is represented as a tuple $(S, A, P, p_0, R)$, where $S$ is the space of possible states of the environment, $A$ is the space of available actions to the agent, $P(s_{t+1}|s_t, a_t)$ is the transition function specifying how the state at time $t+1$ is determined by the current state and action at time $t$, $p_0$ represents the initial state distribution, and $R(s_t, a_t)$ defines the reward received by the agent for a specific state-action pair.
 At each time step of an episode lasting $T$ time steps, the agent observes the environment's state $s_t$, takes an action $a_t$, and receives a reward $r_t$.
 The environment then transitions to the next step according to $P(s_{t+1}|s_t, a_t)$.
 The objective of RL is to learn a policy $\pi_\theta(a|s)$ that maps observed states to actions in order to maximize the cumulative discounted reward $G$ over time, where $G=\sum_{t=0}^{T} \gamma^t r_t$ \cite{sutton2018reinforcement}. The parameter $\gamma < 1$ discounts future rewards during decision making. 
 
 In Deep RL \cite{li2017deep}, the policy is implemented as an artificial neural network, whose connection weights are iteratively updated as the agent interacts with the environment. In all conducted experiments, we employ the Proximal Policy Optimization (PPO) RL algorithm \cite{schulman2017proximal} (see details in Section \ref{ppo}).

 \subsection{Meta Reinforcement Learning as a model of the interplay between evolution and development} \label{meta-RL}

  While RL has led to impressive applications \cite{mnih2013playing,silver2017mastering,berner2019dota}, it suffers from several limitations: the learned policy is specific to the task at hand and does not necessarily generalize well to variations of the environment while requiring a large amount of data to converge.
  To address these issues Meta Reinforcement Learning (Meta-RL) \cite{beck2023survey} aims at training agents that \emph{learn how to learn}, i.e. agents that can quickly adapt to new tasks or environments unknown during training.
  It is based on two nested adaptive loops : an outer loop, analogous to evolution, optimizes the HPs of an inner loop, analogous to development (Fig.~\ref{methods_fig}.b)\cite{pedersen2021evolving,pedersen2023learning}.
  The objective of the outer loop is to maximize the average performance of the inner loop on a distribution of environments. Formally, a set of HPs $\Phi$ are meta-optimized in the outer loop, with the objective of maximizing the average performance of a population of RL agents conditioned by $\Phi$.
  In this paper, we leverage the RC framework where $\Phi$ corresponds to HPs encoding macro-level properties of a RNN, as explained in the next subsection.

  
 \subsection{Reservoir computing as a model of neural structure generation} \label{RC}

Meta-RL algorithms often directly optimize the weights of a RNN through backpropagation in the outer loop \cite{finn2017model,duan2016rl}.
While this technique has demonstrated remarkable efficacy, it is ill-suited for addressing the research question outlined in the introduction.
This is due to its lack of biological plausibility in two main aspects: (1) evolutionary-scale adaptation cannot rely on backpropagation mechanisms \cite{stork1989backpropagation} and (2) the notion that evolution directly fine-tunes neural network weights contradicts the genomic bottleneck principle mentioned in the introduction~\cite{zador2019critique}.
Instead our method evolves RNNs based on the Reservoir Computing (RC) paradigm.
Instead of directly optimizing the neural network weights at the evolutionary scale, it optimizes HPs encoding macro-level properties of randomly generated recurrent networks. 

The fundamental idea behind RC is to create a dynamic 'reservoir' of computation
, where inputs are nonlinearly and recurrently recombined over time~\cite{lukovsevivcius2009reservoir}. This provides a set of dynamic features from which a linear 'readout' can be easily trained: such training equivalent to selecting and combining interesting features to solve the given task (Fig.~\ref{methods_fig}.c). 


A reservoir is generated from a few HPs which play a crucial role in shaping the efficiency of the reservoir dynamics. This includes the number of neurons in the reservoir, the spectral radius $sr$ (controlling the level of recurrence in the generated network), input scaling $iss$ (controlling the strength of the network's inputs), and leak rate $lr$ (controlling how much the neurons retain past information); we explain reservoir HPs in more details in Appendix~\ref{res_hps}. 
In this paper, we propose to meta-optimize reservoir's HPs $\Phi = (sr, iss, lr)$ in a Meta-RL outer loop, using evolutionary algorithms explained in the next subsection. We will then explain how we propose to integrate RC with RL in Section \ref{our_method}. 

 \subsection{Evolutionary algorithms as a model of evolution} \label{EAs}

Evolutionary Algorithms (EAs) draw inspiration from the fundamental principles of biological evolution \cite{back1993overview,reddy2012computational}, where species improve their fitness through the selection and variation of their genomes. EAs iteratively enhance a population of candidate parameterized solutions to a given optimization problem, iteratively selecting those with higher fitness levels (i.e higher performance of the solution) and mutating their parameters for the next generation.

In our approach, we utilize the Covariance Matrix Adaptation Evolution Strategy (CMA-ES) \cite{hansen2016cma} as our designated evolutionary algorithm in order to meta-optimize HPs $\Phi$ of reservoirs. In CMA-ES, a population of HPs candidates is sampled from a multivariate Gaussian distribution, with mean $\mu$ and covariance matrix $V$. The fitness of each sample $\Phi_i$ of the population is evaluated (see Section~\ref{our_method} for how we do it in our proposed method). The Gaussian distribution is then updated by weighting each sample proportionally to its fitness; resulting in a new mean and covariance matrix that are biased toward solutions with higher fitness. This process continues iteratively until either convergence towards sufficiently high fitness values of the generated HPs is achieved, or until a predefined threshold of candidates is reached.

\section{Evolving Reservoirs for Meta Reinforcement Learning (ER-MRL)} \label{our_method}

\subsubsection{General approach}

\begin{figure}[!htbp]
\begin{center}
\includegraphics[width=0.9\textwidth]{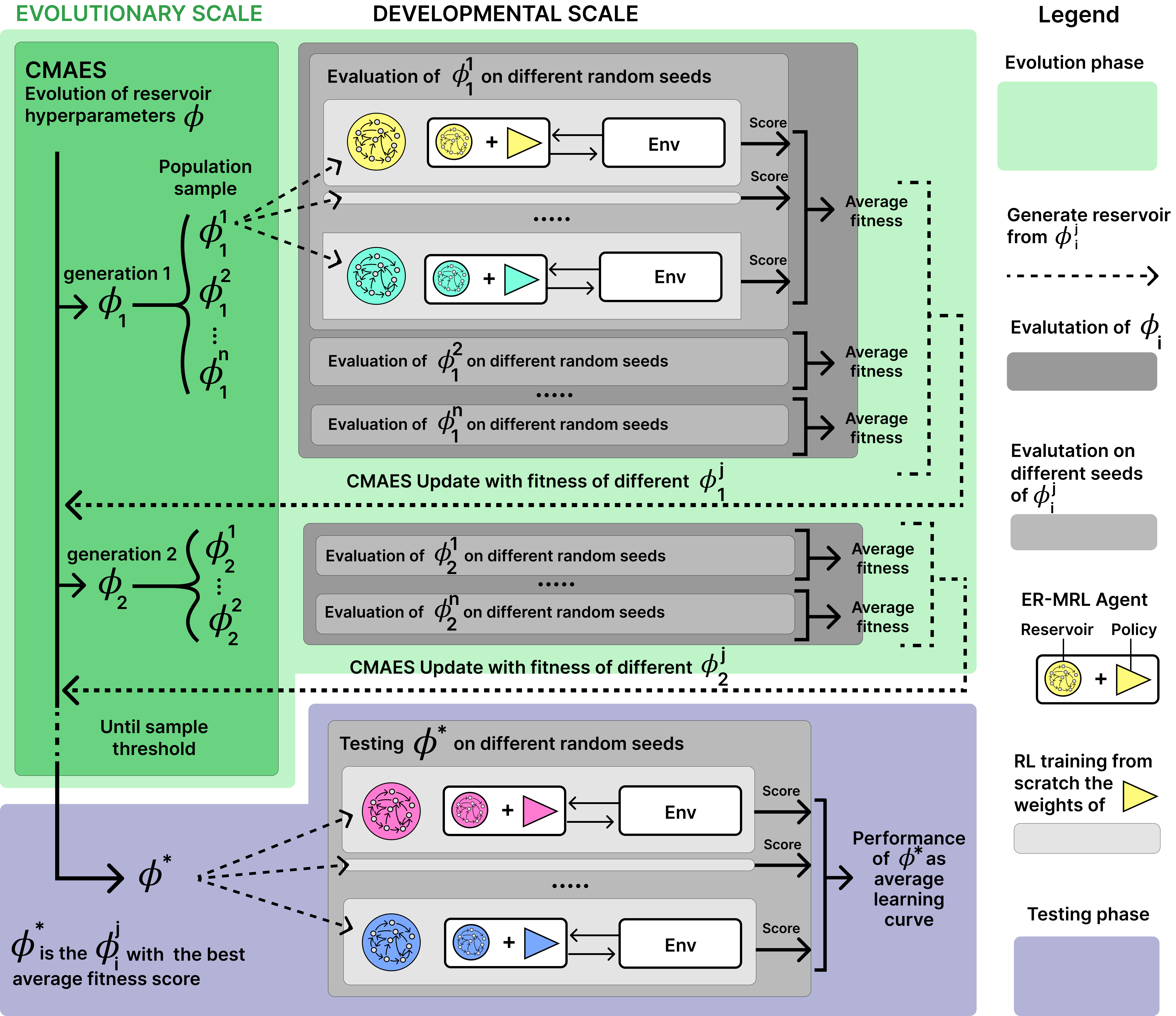}
\caption{In the evolution phase (top),  CMA-ES refines Reservoir HPs $\Phi$. At each generation $i$ of the evolution loop (left), a population $\Phi_i : \{\Phi_i^1, \ldots, \Phi_i^n\}$ of HPs is sampled from the CMA-ES Gaussian distribution. Each $\Phi_i^j$ undergoes evaluation on multiple random seeds, generating multiple reservoirs. An ER-MRL agent is created for each reservoir, with its action policy being trained from the states of that reservoir (lighter grey frames). The fitness of a sampled $\Phi_i^j$ is determined by the average score of all ER-MRL agents generated from it (mid-grey frames). The fitness values are used to update the CMA-ES distribution for the next generation (dotted arrow). This process iterates until a predetermined threshold is reached. In the Testing phase (bottom), the best set of HPs $\Phi^{*}$ from all CMA-ES samples is employed. Multiple reservoirs are generated within ER-MRL agents, and their performance is evaluated.}
\end{center}
\label{details_fig}
\end{figure}

Our objective is to devise a computational framework to address a fundamental question: How can neural structures adapt at an evolutionary scale, enabling agents to better adapt to their environment at a developmental scale? For this aim, we aim to integrate the Machine Learning paradigms presented above. The architecture is illustrated in Fig.~\ref{methods_fig}.d and the optimization procedure in Fig.~\ref{details_fig}. We call our method ER-MRL, for "Evolving Reservoirs for Meta Reinforcement Learning". 

The ER-MRL method encompasses two nested optimization loops (as in Meta-RL, section~\ref{meta-RL}). In the outer loop, operating at an evolutionary scale, HPs $\Phi$ for generating a reservoir (section~\ref{RC}) are optimized  using an evolutionary algorithm (section~\ref{EAs}). In the inner loop, focused on a developmental scale, a RL algorithm (Section~\ref{RL}) learns an action policy $\pi_\theta$ using the reservoir state as inputs. In other words, the outer loop meta-learns HPs able to generate reservoirs resulting in maximal averages performance on multiple inner loops. The whole process is illustrated in Fig.~\ref{details_fig} and detailed below.

\subsubsection{Inner loop}

To represent the development of an agent, we consider a RL agent (Section~\ref{RL}) that interacts with an environment through observation $o_t$, actions $a_t$ and rewards $r_t$ at each time step $t$ (Fig.~\ref{methods_fig}.a). In our proposed ER-MRL method, this agent is composed of three main parts : a reservoir generated by HPs $\Phi=\{iss,lr,sr\}$ (see Section~\ref{res_hps} for more details), a feed forward action policy network $\pi_\theta$ and a RL algorithm. At each time step, we feed the reservoir with the current $o_t$, and the previous action and reward $a_t$ and $r_t$ (Fig.~\ref{methods_fig}.d). Contrarily to standard RL, the policy $\pi_\theta$ does not directly access the observation of the environment's state $o_t$, but the context $c_t$ of the reservoir instead (i.e. the vector of all reservoir's neurons activations at time $t$). Because reservoirs are recurrent neural networks, $c_t$ not only encompasses information about the current time step, but also integrates information over previous time steps. In some experiments, we also use ER-MRL with multiple reservoirs. In this case, we still generate the reservoirs from a set of HPs $\Phi$, and the context $c_t$ given to the policy is the concatenation of hidden states of all reservoirs. We then train our policy $\pi_\theta(a|c_t)$ using RL. 



\subsubsection{Outer loop}

The outer loop employs the Covariance Matrix Adaptation Evolutionary Strategy (CMA-ES) (Section~\ref{EAs}) to optimize reservoir HPs $\Phi$. The objective is to generate reservoirs which, on average over multiple agents, improve learning abilities. For each set of HPs, we assess the performance of our agents in multiple inner loops (we utilize 3 in our experiments), each one with a different random seed. Using different random seeds implies that, while using the same HPs set, each agent will be initialized with different connection weights of both their reservoirs, their policies and the initial environment state. Note that while the generated reservoirs have different connection weights, they share the same macro-properties in terms of spectral radius $sr$, input scaling $iss$ and leak rate $lr$ (since they are generated from the same HPs set). In assessing an agent's fitness within its RL environment, we compute the mean episodic reward over the final 10 episodes of its training. To obtain the fitness of a reservoir HPs, we calculate the mean fitness of three agents across three different versions of the same environment. These steps are iterated until we reach a predetermined threshold of CMA-ES iterations (set at 900 in our experiments).

\subsubsection{Evaluation}

To evaluate our method, we select the HPs $\Phi^{*}$ that generated the best fitness function during the whole outer loop optimization with CMA-ES (see bottom of Fig.~\ref{details_fig}). We then generate 10 ER-MRL agents with different random seeds (with a different reservoir sampled from $\Phi^{*}$ for each seed, together with random initial policy weights $\theta$) and train the action policy $\pi_\theta$ of each agent using RL. We report our results in the next section, comparing the performance of these agents against vanilla RL agents using a feedfoward policy.


 \section{Results} \label{results}

We designed experiments to study the following hypotheses : The ER-MRL architecture combining reservoirs and RL could enable (1) solving tasks with partial observability, (2) generating oscillatory dynamics that facilitate the learning of locomotion tasks, and (3) facilitating the generalization of learned behaviors to new tasks unseen during evolution phase.

 \subsection{Evolved reservoirs improve learning in highly partially observable environments}

In this section, we evaluate our approach on tasks with partial observability, where we purposefully remove information from the agent observations. Our hypothesis is that the evolved reservoir can help reconstructing this missing information. Partial observability is an important challenge in the field of RL, where agents have access to only a limited portion of environmental information to make decisions. This is referred to as a Partially Observable Markov Decision Process (POMDP) \cite{monahan1982state} rather than a traditional MDP.  In this context, the task becomes harder to learn, or even impossible, as the agent needs to make decisions based on an incomplete observation of the environment state. To explore this issue, our experimental framework is based on control environments, such as CartPole, Pendulum, and LunarLander (see details in Fig.~\ref{part_obs_envs_fig} of the appendix). We modify these environments by removing velocity-related observations, thus simulating a partially-observable task.


Let's illustrate this issue with the first environment (CartPole), where the agent's goal is to keep the pole upright on the cart while it moves laterally. If we remove velocity-related observations (both for the cart and the pole's angle), a standard feedfoward RL agent cannot effectively solve the task. The reason is straightforward: without this information, the agent doesn't know the cart's movement direction or whether the pole is falling or rising. We apply the same process to the other two environments, removing all velocity-related observations for our agents. Can the ER-MRL architecture address this challenge? To find out, we independently evolve reservoirs using ER-MRL for each task. We search for effective HPs tailored to the partial observability of each environment. To evaluate our approach, we compare the learning curves of ER-MRL agents (from the test phase, see bottom of Fig.~\ref{details_fig}) on these three partially observable environments against an agent with a feedforward policy. 

\begin{figure}
\begin{center}
\includegraphics[width=0.9\textwidth]{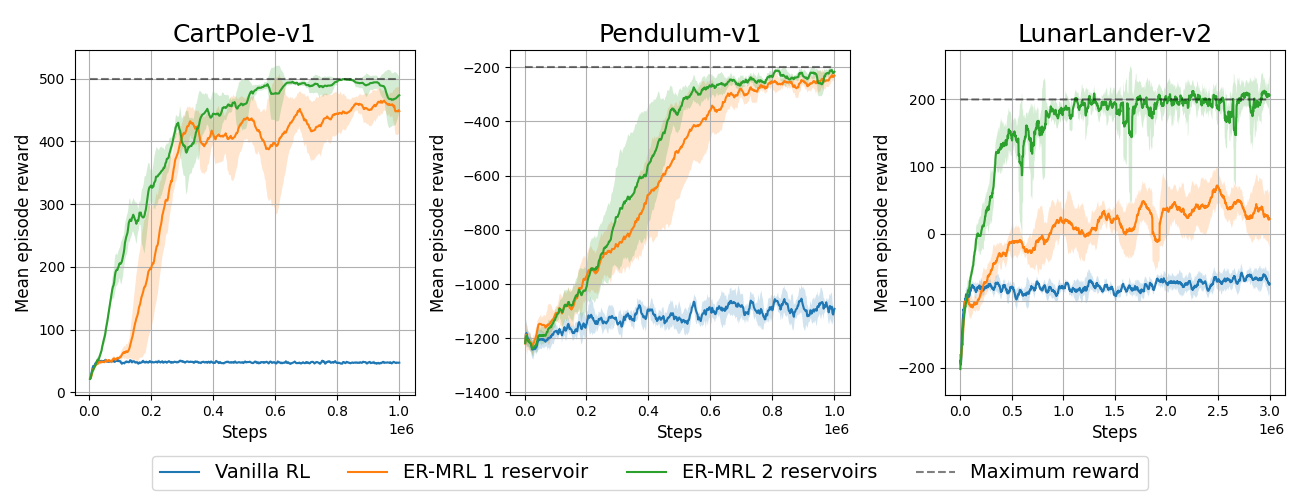}
\caption{Learning curves for partially observable tasks. The x-axis represents the number of timesteps during the training and the y-axis the mean episodic reward. Learning curves of our ER-MRL methods correspond to the testing phase described in the bottom of Fig.~\ref{details_fig}. Vanilla RL corresponds to a feedforward policy RL agent. The curves and the shaded areas represent the mean and the standard deviation of the reward for 10 random seeds. See Section~\ref{part_obs_benchmark} for a comparison with another method.} 
\end{center}
\label{po_fig}
\end{figure}

Fig.~\ref{po_fig} presents the results for the three selected partially observable tasks. We observe, as expected, that vanilla RL agents cannot learn how to solve the task under partial observability (for the reasons mentioned above). In comparison, our approach leads to performance scores close to those obtained by a RL algorithm with full observability. This indicates that the evolved reservoir is able to reconstruct missing information related to velocities from its own internal recurrent dynamics. This confirms the hypothesis that an agent with a reservoir can solve partially observable tasks by using the internal reservoir state to reconstruct missing information. We explain with more details why this method could work in Section~\ref{part_obs_results_expl}  of the appendix. The difference in results between the model with 2 reservoirs on LunarLander environment suggests that solving it requires encoding at least two different timescales dynamics. Our interpretation here is that solving LunarLander requires to deal with both an "approaching" and "landing" phase, unlike the two other environments. 

 \subsection{Evolved reservoirs could generate oscillatory dynamics that facilitate the learning of locomotion tasks}


 In this section, we evaluate our approach on agents with 3D morphology having to learn locomotion tasks shown in Fig.~\ref{mujoco_envs_fig}. We postulate that the integration of an evolved reservoir can engender oscillatory patterns that aid in coordinating body movements, akin to Central Pattern Generators (CPGs). CPGs, rooted in neurobiology, denote an interconnected network of neurons responsible for generating intricate and repetitive rhythmic patterns that govern movements or behaviors \cite{marder2001central} such as walking, swimming, or other cyclical movements. Existing scientific literature hypothesizes that reservoirs, possessing significant rhythmic components, share direct connections with CPGs \cite{ren2015multiple}. We propose to study this hypothesis using motor tasks involving rhythmic movements.
 
 We employed 3D MuJoCo environments (detailed in Fig.~\ref{mujoco_envs_fig} of the appendix), where the goal is to exert forces on various rotors of creatures to propel them forward. Notably, while the ultimate goal across these tasks remains constant (forward movement), the creatures exhibit diverse morphologies, including humanoids, insects, worms, bipeds, and more. Furthermore, the action and observation spaces vary for each morphology. We individually evaluate our ER-MRL architecture on each of these tasks. 
 
 

\begin{figure}
\begin{center}
\includegraphics[width=0.9\textwidth]{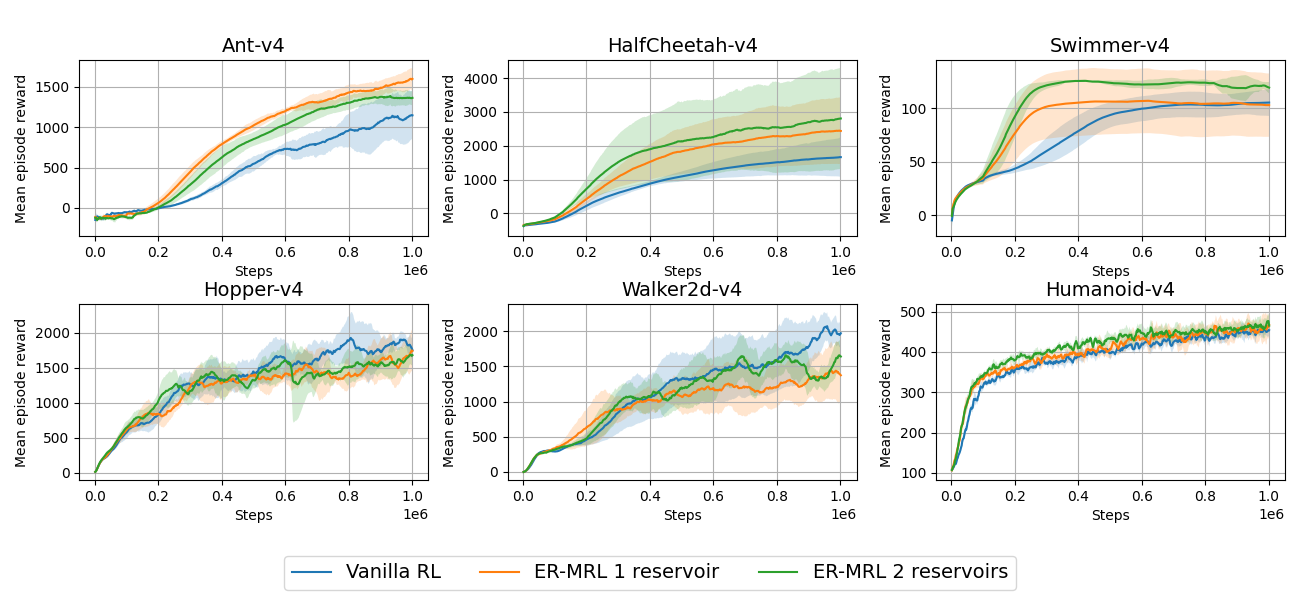}
\caption{Learning curves for locomotion tasks. Same conventions as Fig.~\ref{po_fig}} \label{mujoco_fig}
\end{center}
\end{figure}

Our approach demonstrates improved performance in some tasks (Ant, HalfCheetah, and Swimmer) compared to a standard RL baseline, particularly noticeable in the early stages of learning, as illustrated in Fig.~\ref{mujoco_fig}. This suggests that the evolved reservoir may generate beneficial oscillatory patterns, facilitating the learning of locomotion tasks, in line with the notion that reservoirs could potentially function as CPGs, aiding in solving motor tasks. Although carefully testing this hypotheses would require more analysis, we present in Section~\ref{res_cpgs} in the appendix preliminary data suggesting that the evolved reservoir is able to generate  oscillatory dynamics that could facilitate learning in the Swimmer environment. However, as shown in Fig.~\ref{mujoco_fig}, performance enhancement was not observed in the Walker and Hopper environments compared to the RL baseline. Locomotion in both environments demands precise closed-loop control strategies to maintain an agent's equilibrium. In such cases, generated oscillatory patterns may not be as beneficial. 


 \subsection{Evolved reservoirs improve generalization on new tasks unseen during evolution phase} \label{generalization_section}


 In this section, we address a key aspect of our study: the ability of evolved reservoirs to facilitate adaptation to novel environments. This inquiry is crucial in assessing the potential of evolved neural structures to generalize and enhance an agent's adaptability beyond the evolution phase. Building on the promising results of ER-MRL with two reservoirs in previous experiments, we focus exclusively on this configuration for comparison with the RL baseline.
 
\subsubsection{Generalizing across different morphologies with similar tasks.} \label{gener_diff_morpho}


In prior experiments, ER-MRL demonstrated effectiveness in environments like Ant, HalfCheetah, and Swimmer. This success led us to explore whether reservoirs evolved for two of these tasks could be adaptable to the third, indicating potential generalization across different morphologies. However, due to variations in environments, including differences in morphology, observation and action spaces, and reward functions, generalization from one set of tasks to another presents a complex challenge. To ensure fair task representation of each environment in the final fitness, we employ the normalization formula detailed in Section~\ref{normalized_scores}. Subsequently, we select the reservoir HPs $\Phi^{*}$ that yielded the highest fitness and evaluate them in a distinct environment. For instance, if we evolve reservoirs on Ant and HalfCheetah, we test them in the Swimmer task.

\begin{figure}
\begin{center}
\includegraphics[width=0.9\textwidth]{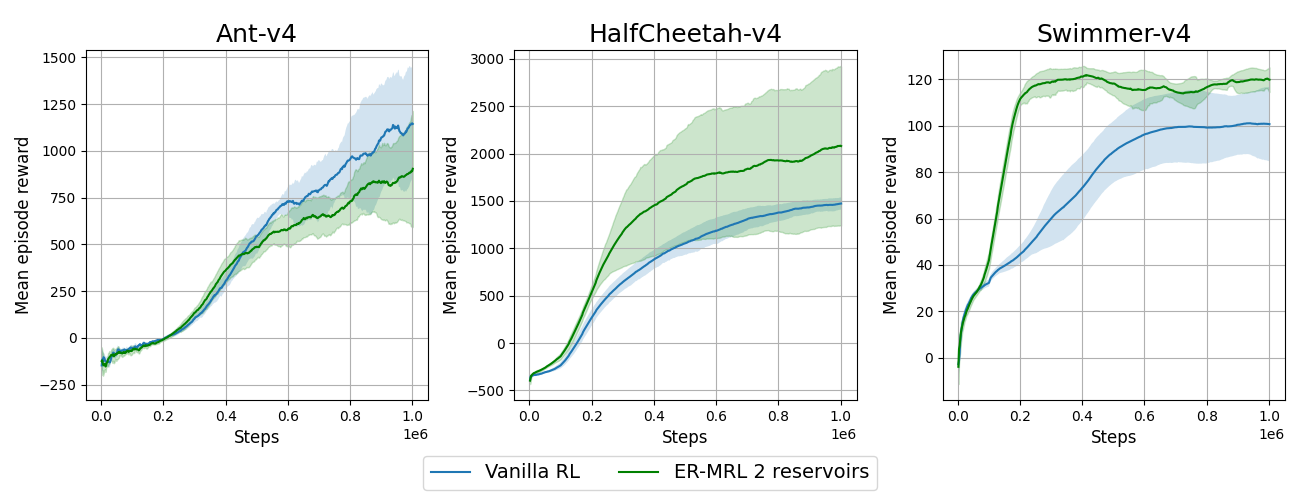}
\caption{Learning curves for generalization on similar locomotion tasks with different morphologies. The curves evaluate the performance of ER-MRL on an environment that was unseen during the evolution phase. For instance, the left plot shows performance of an agent on Ant, using reservoirs evolved on only HalfCheetah and Swimmer.} \label{generalization-2-envs}
\end{center}
\end{figure}


In Fig.~\ref{generalization-2-envs}, we observed a notable improvement in the performance of ER-MRL agents with reservoirs evolved for different tasks, particularly in HalfCheetah and Swimmer environments. This substantiates the capacity of evolved reservoirs to generalize to new tasks and encode diverse dynamics from environments with distinct morphologies. However, it's worth noting that this improvement wasn't replicated in the Ant task. This could be attributed to the unique characteristics of the Ant environment, with its stable four legged structure, in contrast to the simpler anatomies of Swimmer and HalfCheetah. For a detailed analysis, please refer to Section~\ref{hps_analysis} in the appendix.

\begin{figure}
\begin{center}
\includegraphics[width=0.9\textwidth]{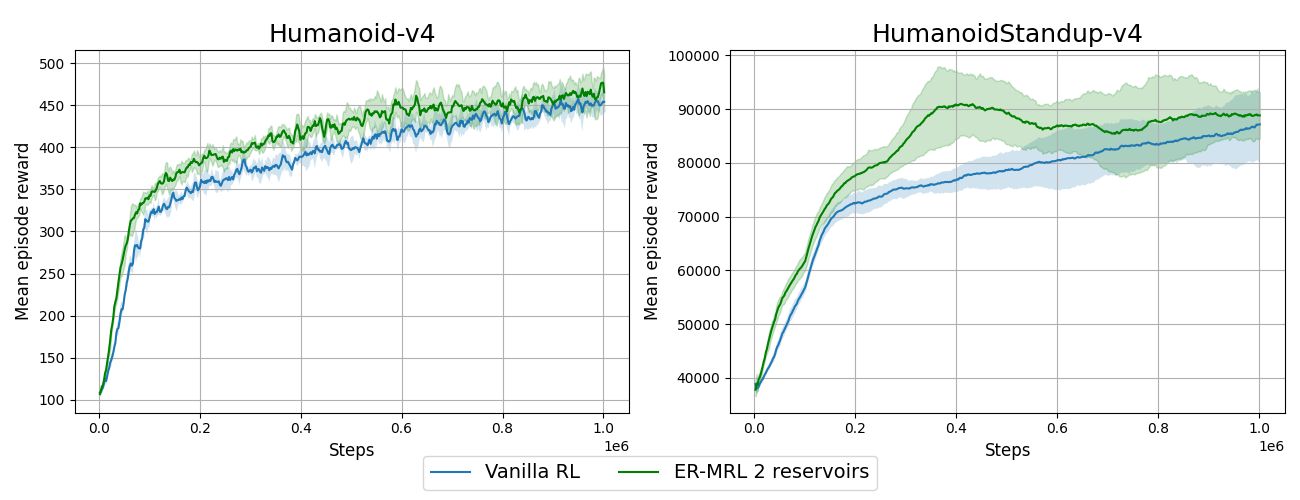}
\caption{Learning curves for generalization on different locomotion tasks with similar morphologies. The reservoirs are evolved on one task and tested on the other one.} \label{humanoids_results}
\end{center}
\end{figure}

\subsubsection{Generalizing across different tasks with similar morphologies.} \label{gener_humanoids}

We have seen how reservoirs facilitated ER-MRL agent's ability to generalize across locomotion tasks with different morphologies. Now, we shift our focus to tasks with consistent morphologies but distinct objectives. To delve into this, we turn to the Humanoid and HumanoidStandup environments (shown in Fig.~\ref{humanoids_envs} of the appendix), both presenting tasks within the realm of humanoid movement. One task involves learning to walk as far as possible, while the other centers around the challenge of standing up from the ground. As in our previous study, we follow the procedure of evolving reservoir-generating HPs on one task and evaluating their performance on the other.

Fig.~\ref{humanoids_results} provides a visual representation of our findings. While the performance improvement may not be dramatic, it underscores the generalization capabilities of reservoirs across tasks with similar morphologies but differing objectives. This observation, though promising, invites further investigation, given the limited number of experiments conducted in this context. This aspect represents an avenue for future research.

\section{Discussion}

In this paper, we have addressed the compelling question of whether reservoir-like neural structures can be evolved at an evolutionary time scale, to facilitate the learning of agents on a multitude of sensorimotor tasks at a developmental scale. Our results demonstrate the effectiveness of employing evolutionary algorithms to optimize these reservoirs, especially on Reinforcement Learning tasks involving partial observability, locomotion, and generalization of evolved reservoirs to unseen tasks. 

Our ER-MRL approach has parallels to previous algorithms in RL that employ an indirect encoding for mapping a genome to a particular neural network architecture~\cite{ha_hypernetworks_2016,stanley_hypercube-based_2009,najarro2023towards} . Our choice of employing reservoirs comes with the benefit of a very small genomic size (reservoirs are parameterised by a handful of parameters that we show in Appendix ~\ref{res_hps}) without reducing the complexity of the phenotype (the number of weights of the reservoir policy is independent of the number of hyper-parameters). Moreover, our approach clearly distinguish neural structures optimized at the evolutionary scale (the reservoirs) vs. at the developmental scale (the RL action policy).

Nonetheless, some limitations persist within our methodology. The combination of reservoir computing and reinforcement learning remains underexplored in the existing literature \cite{chang2020reinforcement,chang2018distributive}, leaving substantial room for refining the algorithmic framework for improved performance. Moreover, our generalization experiments and quantitative analyses warrant further extensive testing to gain deeper insights. Notably, our approach does incur a computational cost due to the time required to train a new policy with RL for each generated reservoir. Future studies could devise more efficient evolutionary strategies or employ alternative optimization techniques.

However, because our method remains agnostic to specific environment and agent's characteristics (a reservoir architecture being independent of the shape of its inputs and outputs), we could in theory evolve reservoirs across a very wide range of environments and agent's morphologies. Such evolved generalist reservoirs could then result in highly reduced computational cost at the developmental scale, as our results suggest, compared to training recurrent architectures from scratch.  

Moving forward, there are several promising avenues for exploration. Firstly, a more comprehensive understanding of the interaction between RL and RC could significantly improve the performance of such methods on developmental learning tasks. Secondly, integrating our approach with more sophisticated Meta-RL algorithms could offer a mean to initialize RL policy weights with purposefully selected values rather than random ones. Additionally, a broader framework allowing for the evolution of neural structures with greater flexibility, such as varying HPs and neuron counts, could yield more intricate patterns during the evolution phase, potentially resulting in substantial improvements in agent performance across developmental tasks \cite{stanley2009hypercube,najarro2023towards}.

Our research bridges the gap between evolutionary algorithms, reservoir computing and meta-reinforcement learning, creating a robust framework for modelling neural architecture evolution. We believe that this integrative approach opens up exciting perspectives for future research in RC and Meta-RL to propose new paradigms of computations. It also provides a computational framework to study the complex interplay between evolution and development, a central issue in modern biology \cite{johnston1982selective,hougen2019evolution,watson2016can,moulin2022ecology}.



\section*{Acknowledgments}
Financial support was received from: the University of Bordeaux's France 2030 program / RRI PHDS framework, French National Research Agency (ANR) grants: ECOCURL ANR-20-CE23-0006 and DEEPPOOL ANR-21-CE23-0009-01. We benefited HPC resources : IDRIS under the allocation A0091011996 made by GENCI, using the Jean Zay supercomputer, and Curta from the University of Bordeaux.

%
%
%
\bibliographystyle{splncs04}
\bibliography{references}

\section{Appendix}

In this appendix, we provide comprehensive insights and clarifications on the methodologies employed in our study. Specifically, we elaborate on aspects such as the parameters governing our experiments, including the RL (PPO), the RC and the evolutionary (CMA-ES) algorithms we used. Furthermore, we furnish a detailed exposition of the environments utilized in our research. Lastly, we conduct supplementary analyses aimed at enhancing our understanding of some observed phenomena in the obtained results. In addition, we present results from experiments that were not featured in the main text to offer a more comprehensive view of our findings.

\subsection{Methods}

\subsubsection{Proximal Policy Optimization (PPO)} \label{ppo}

 PPO, categorized as a policy gradient technique \cite{sutton1999policy}, undertakes exploration of diverse policies through stochastic gradient ascent. This process involves assigning elevated probabilities to actions correlated with high rewards, subsequently adjusting the policy to aim for higher expected returns. The adoption of PPO stems from its well-established reputation as a highly efficient and stable algorithm in the scientific literature, although its use does not have major theoretical implications for this particular project. 

 \subsubsection{Reservoir hyperparameters} \label{res_hps}

In Reservoir Computing, the spectral radius controls the trade-off between stability and chaoticity of reservoir dynamics: in general \enquote{edge of chaos} dynamics are often desired \cite{bertschinger2004real}. Input scaling determines the strength of input signals, and the leak rate governs the memory capacity of reservoir neurons over time. These HPs specify the generation of the reservoir weights. Once the reservoir is generated, its weights are kept fixed and only a readout layer, mapping the states of the reservoir neurons to the desired output of the network are learned. Other HPs exist to initialize a reservoir, but they have not been studied in the experiments of the paper (as it has been tested that they have much less influence on the results).

\begin{figure}
\includegraphics[width=0.9\textwidth]{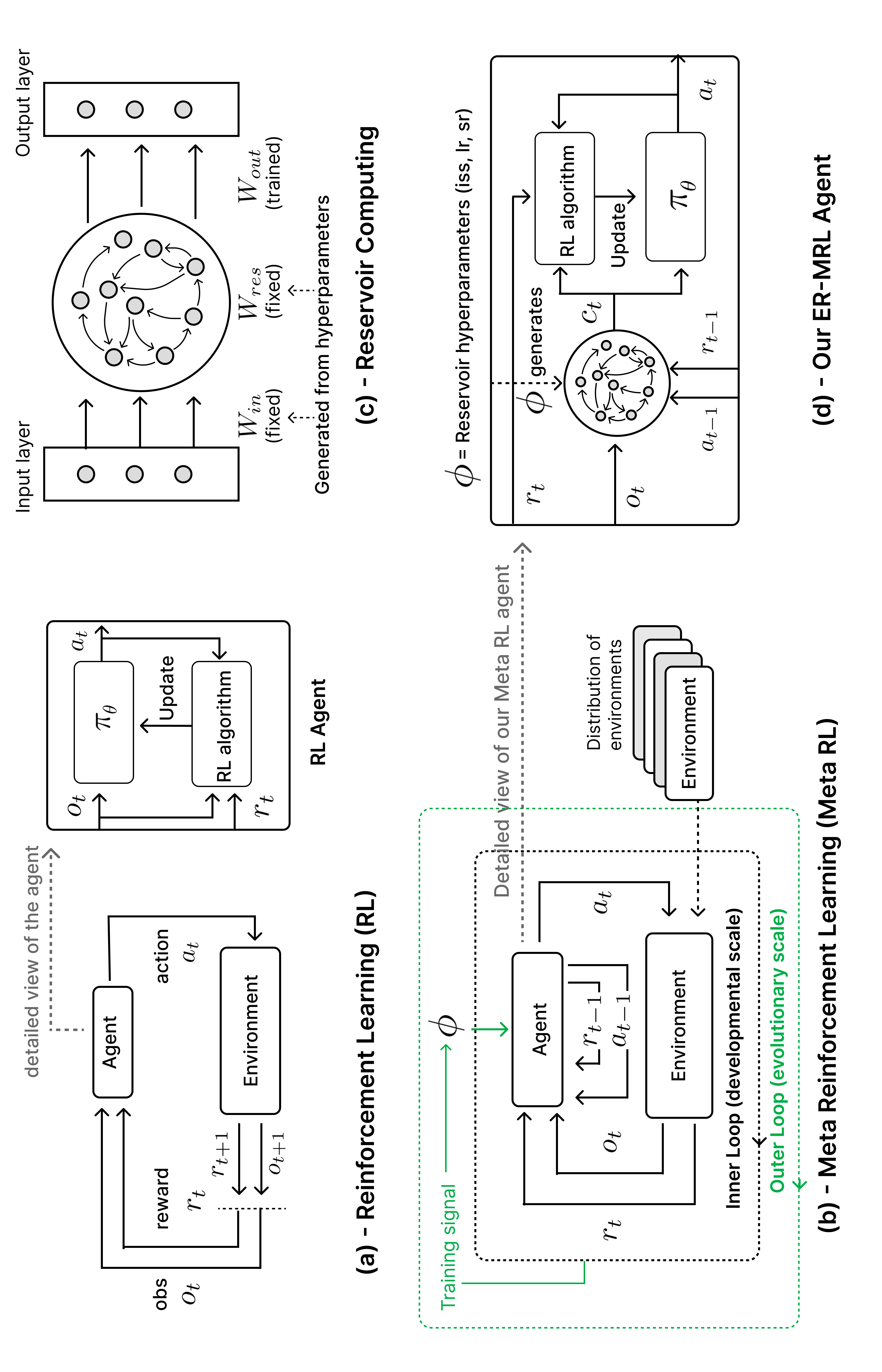}
\caption{Rotated view of Fig.~\ref{methods_fig} presenting the background methods used, and how our ER-MRL agents incorporate them} \label{rotated_methods_fig}
\end{figure}

\subsection{Experiment Parameters}

\subsubsection{General parameters}

In our experiments, we adapted the number of timesteps during the training phase of our ER-MRL agent in the inner loop, based on whether we were evolving the reservoir HPs or testing the best HPs set discovered during the CMA-ES evolution. For the evolution phase, which was computationally intensive, we utilized 300,000 timesteps per training. Conversely, when evaluating our agents against standard RL agents, we employed 1,000,000 timesteps. Notably, in the case of the LunarLander environment, we extended the testing to 3,000,000 timesteps, as the learning curve had not yet converged at 1,000,000 timesteps.

\subsubsection{PPO hyperparameters}

Regarding the parameters of our RL algorithm, PPO, we used \href{https://stable-baselines3.readthedocs.io/en/master/modules/ppo.html}{the default settings} provided by the Stable Baselines3 library \cite{raffin2021stable}. For tasks involving partial observability, we made a slight adjustment by setting the learning rate to 0.0001, as opposed to the standard 0.0003. This modification notably enhanced performance, potentially indicating that reservoirs contained a degree of noise, warranting a lower learning rate to stabilize RL training.

\subsubsection{CMA-ES hyperparameters}

For the parameters of our evolutionary algorithm, CMA-ES, we adopted the default settings of the \href{https://optuna.readthedocs.io/en/v2.10.1/reference/generated/optuna.samplers.CmaEsSampler.html}{CMA-ES sampler} from the Optuna library \cite{akiba2019optuna}.

\subsubsection{Reservoirs hyperparameters}

For the reservoirs, we only modified the parameters mentioned in \ref{res_hps} and the number of neurons. We consistently used 100 neurons per reservoirs during all experiments. All the other HPs were kept the same and are the default \href{https://reservoirpy.readthedocs.io/en/latest/user_guide/hyper.html}{reservoir parameters} used in ReservoirPy \cite{trouvain2020reservoirpy}. We conducted additional analyses and observed that they exerted a relatively modest influence on tasks of this nature. However, a more refined analysis of the importance of these HPs could be interesting in future works.

\subsection{Partially observable environments}

In the following section, we present the different Reinforcement Learning environments from the Gymnasium library \cite{towers_gymnasium_2023}, used during our experiments on partial observability.


 \begin{figure}[!htb]
\includegraphics[width=0.9\textwidth]{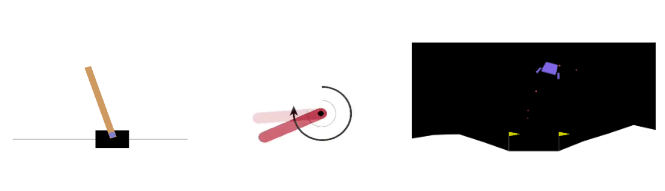}
\caption{Partially observable environments used, The goal of CartPole (left) is to learn how to balance the pole on the cart. The goal of Pendulum (middle) is to learn how to maintain the pendulum straight up by applying forces on it. The goal of LunarLander (right) is to learn how to land between the two flags by generating forces on the different spaceship reactors.} \label{part_obs_envs_fig}
\end{figure}



\subsubsection{Results analysis} \label{part_obs_results_expl}

To better understand the reservoir's capabilities on these tasks, we conducted several tests on supervised learning problems where a sequence of actions, rewards, and observations (without velocity) was provided to a reservoir with a linear readout. In one case, the model had to reconstruct full observation information (position, angle, velocity, angular velocity), and in the other, it had to reconstruct positions and angles over several time steps (doing this only for the last 2 time steps allows a PPO to achieve maximum reward later on). In both cases, this model successfully solved the tasks with very high performance. Moreover, it was also capable of predicting future observations, which can be extremely valuable to find an optimal action policy.

\subsubsection{Benchmark comparison} \label{part_obs_benchmark}

Regarding benchmarks, our approach compares favorably with the results reported in \href{https://wandb.ai/sb3/no-vel-envs/reports/PPO-vs-RecurrentPPO-aka-PPO-LSTM-on-environments-with-masked-velocity--VmlldzoxOTI4NjE4}{the blog post from Raffin}~\cite{raffin2023} where he used another model combining a RNN (LSTM \cite{yu2019review}) with a RL algorithm (PPO, the one we also used) on the same partially observable tasks. The performance on each environment are pretty similar, but it is the training timesteps needed to reach the maximum performance that varies the most between the methods. Indeed for the LunarLander environment, our method is able to learn in less timesteps after evolving reservoirs, but it is the contrary with CartPole and Pendulum tasks. 

It is worth noting that even if both approaches have similarities, ER-MRL consists in optimizing the HPs of reservoirs at an evolutionary scale, whereas the method presented in the blog post trains a recurrent architecture from scratch. This divergence complicates direct comparisons between both methods. Indeed, our results are derived after an extensive phase of computation in a Meta-RL outer loop, but the subsequent evaluation with the final reservoir configuration is comparatively swift. as only the RL policy (linear readout)  requires training. In contrast, the LSTM-PPO method does not incorporate a computationally intensive meta-learning phase, but their training process takes more time per timestep update. Indeed, each training step of the this demands more computation, due to having to train the LSTM from scratch in addition to PPO, compared to our method where only the linear readout is trained at the developmental scale. 

However to ensure a fair and comprehensive comparison with other baselines, especially in tracking the time required to achieve presented results, more experiments are necessary.

\subsection{MuJoCo forward locomotion environments}

 \begin{figure}[!htb]
\includegraphics[width=0.9\textwidth]{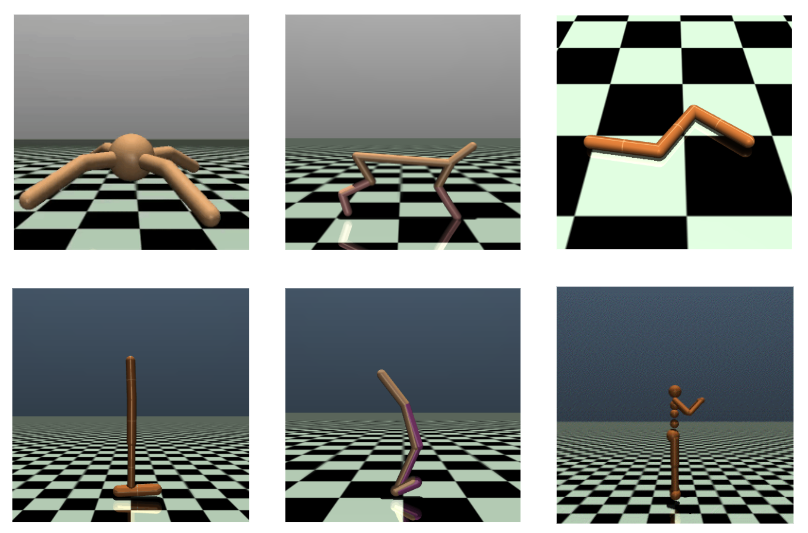}
\caption{MuJoCo environments, the goal of these tasks is to apply force to the rotors of the creatures to make them move forward. On the top row, we have from left to right the Ant, HalfCheetah and Swimmer environments, and on the bottom row, the Hopper, Walker and Humanoid environments. The environment observations comprise positional data of distinct body parts of the creatures, followed by the velocities of those individual components, while actions entail the torques applied to the hinge joints.
 } \label{mujoco_envs_fig}
\end{figure}

\subsubsection{Results analysis} \label{res_cpgs}

In this section, we present how reservoirs could act as Central Pattern Generators within agents learning a locomotion task in these 3D environments. 

\begin{figure}[!htb]
\includegraphics[width=0.9\textwidth]{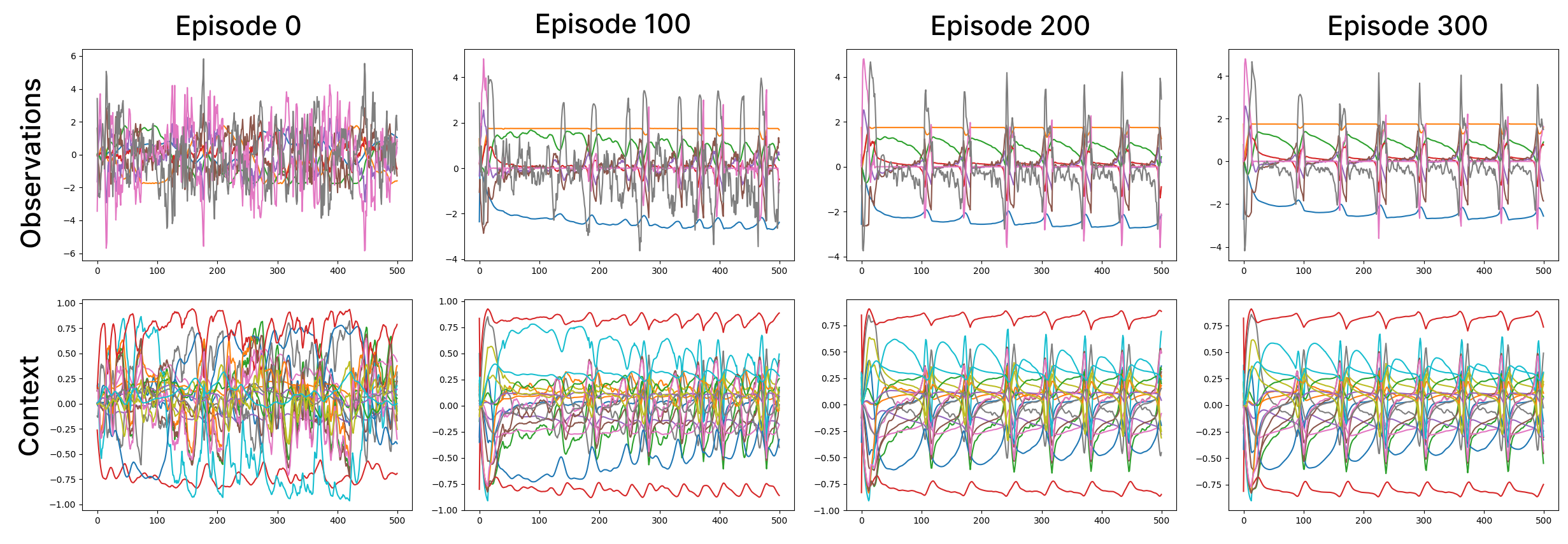}
\caption{Differences between the observations of a RL agent (top) with the context of an ER-MRL agent (bottom) at the same stage of training. Each episode lasts 1000 timesteps in the environment. The curves of the RL agent represent the real observation values from the environment, and the curves of the ER-MRL one part of the context given to the agent's policy : the activation values of 20 reservoir neurons (out of 100).} \label{context-observation}
\end{figure}

It can be observed that the separation between the two models seems to occur starting from 100,000 timesteps at the top-right of Fig.~\ref{mujoco_fig}. Therefore, we recorded \href{https://docs.google.com/presentation/d/1F-p-FKROFdIWIUj0BY-TW2k6buaJskjX8p19S61inqE/edit?usp=sharing}{videos of the RL and ER-MRL agents} to better understand the performance difference between the two models. Furthermore, we conducted a study at the level of the input vector in the agent's policy ($o_t$ for RL agent, and $c_t$ for ER-MRL agent). As seen in Fig.~\ref{context-observation}, it is noticeable that very early in the learning process, the reservoir exhibits much more rhythmic dynamics than the sole observation provided by the environment. This could be due to the link between the reservoir and CPGs, potentially facilitating the acquisition and learning of motor control in these tasks. 

Expanding on this, it's notable that CPGs, shared across various species, have evolved to embody common structures. Drawing parallels from nature, our investigation delves into whether generalization (results in Section ~\ref{generalization_section}) across a spectrum of motor tasks may mirror the principles found in biological systems. 

However, further experiments, accompanied by robust quantitative analysis, are necessary to gain valuable insights into whether reservoirs can function as CPG-like structures.

\subsection{MuJoCo humanoid environments}

 \begin{figure}[!htb]
\includegraphics[width=0.9\textwidth]{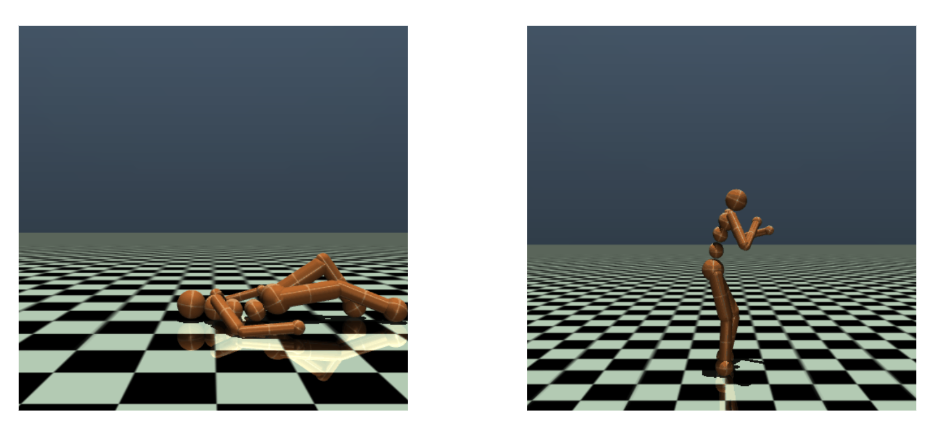}
\caption{MuJoCo environments with humanoid morphologies. On the left figure, the goal is to learn how to stand up, and on the right the goal is to walk forward as far as possible} \label{humanoids_envs}
\end{figure}

\subsubsection{Interesting Reservoir results}

As seen in Section \ref{RC}, one of the basic principles of RC is to project input data into a higher-dimensional space. In the case of the Humanoid tasks, where our results are displayed in Fig.~\ref{mujoco_fig} and Fig.~\ref{humanoids_results}, the initial observation and action space is larger (400 dimensions) compared to the context dimension for one or two reservoirs of 100 neurons (the dimension is equal to the number of neurons) . This means that even by reducing the input dimension in the RL policy network, the reservoir improves the quality of the data. For other morphologies, the dimension of input data is inferior to the dimension our reservoir context.

\subsection{Normalized scores for generalization} \label{normalized_scores}

To prevent any particular task from disproportionately influencing the fitness score due to variations in reward scales, we use a fitness function for CMA-ES that aggregates the normalized score, denoted as $nScore$, across both environments. The normalization process is defined as :

$$nScore = \frac{score - randomScore}{baselineScore - randomScore}$$

\vspace{5mm}

Where $randomScore$ and $baselineScore$ represent the performances of a random and of a standard PPO agent, respectively.

\subsection{Reservoir hyperparameters analysis} \label{hps_analysis}

In preceding sections, we observed how HPs play a pivotal role in enabling ER-MRL agents to generalize across tasks. Now, we delve deeper into understanding why some reservoirs aid in generalization for specific tasks while others do not. To gain this insight, we constructed a hyperparameter map to visualize the regions of HPs associated with the best fitness in each environment. We selected the best 30 sets of HPs, comprising the spectral radius and leak rate values of the reservoirs, out of a pool of 900 for all MuJoCo locomotion tasks (refer to Fig.~\ref{mujoco_envs_fig}) and plotted them on a 2D plane.

\begin{figure}
\includegraphics[width=0.9\textwidth]{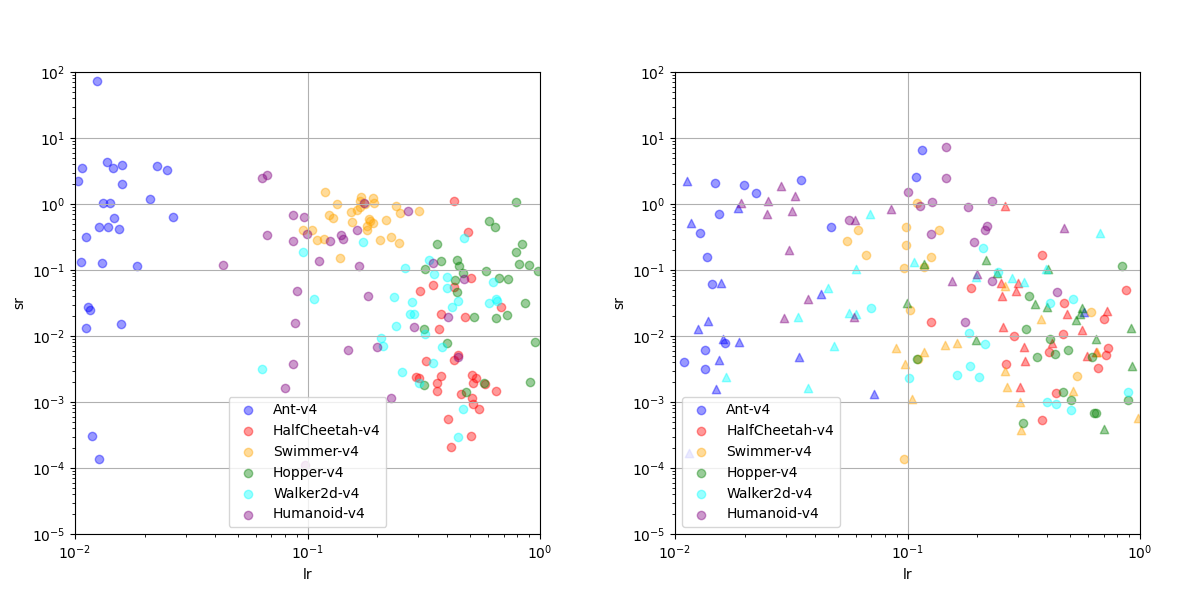}
\caption{The left figure represents parameters obtained with a single reservoir, while the right figure corresponds to configurations with two reservoirs (depicted as either circles or triangles).} \label{hps_map}
\end{figure}

In Fig.~\ref{hps_map}, we observe that the HPs for most environments are clustered closely together. Conversely, those for the Ant environment form a distinct cluster, characterized by notably lower leak rates. The leak rate reflects how much information a neuron retains in the reservoir, influencing its responsiveness to input data and connections with other neurons. A lower leak rate implies a more extended memory, possibly instrumental in capturing long-term dynamics. This observation aligns with the stable morphology of the Ant, potentially allowing the agent to prioritize long-term dynamics for efficient locomotion. This would partially explain why generalization wasn't successful on this environment in Section \ref{gener_diff_morpho}, when reservoirs were evolved on other types of morphologies.

\section{Additional experiments}

We also led other experiments that we didn't mention in the main text. As mentioned above in Section~\ref{res_hps}, we consistently employed reservoirs with a size of 100 neurons to ensure a standardized basis for result comparison. This configuration equates one reservoir to 100 neurons, two reservoirs to 200 neurons, and so forth. We conducted additional experiments to investigate the impact of varying the number of reservoirs and neurons within them. We observed that altering the number of neurons within a reservoir had a limited effect. For example, reducing the number of neurons to as low as 25 did not significantly affect performance on the partially observable environments. Increasing the size of the reservoirs didn't seem to improve the performance a lot either, except for the Humanoid environments (with a large observation space) where reservoirs equipped with a lot of neurons (1000) performed slightly better than others. While we opted for 100 neurons per reservoir in our experiments, there is surely potential for further optimization.

Furthermore, we explored experiments involving partially observable reservoirs, in which only a subset of the observation was provided to the policy. The results demonstrated that it is not always necessary to fully observe the contextual information within the reservoir to successfully accomplish tasks. On the CartPole environment, we tested 3 type of models with a reservoir of 100 fully observable neurons (the policy has access to 100 out of the 100 neurons), a reservoir of 1000 fully observable neurons, and another reservoir with only 100 partially observable neurons out of 1000. We observed that the model with 1000 fully observable neurons performed worse than the two other, who had similar results.

Regarding generalization experiments, we investigated the impact of varying the number of reservoirs. Although experiments with three reservoirs yielded intriguing insights, such as distinct memory types characterized by leak rate in the different reservoirs, the overall performance was notably lower compared to configurations with two reservoirs. This observation can likely be attributed to the increased complexity of learning due to the larger observation space, despite the potential for richer dynamics. We also noted instances where several reservoirs maintained very similar hyperparameters for specific tasks, potentially indicating the importance of capturing particular dynamics.

Additionally, we considered the possibility of employing smaller reservoirs in greater numbers. This approach could capture a diverse range of interesting features, such as different dynamics, while keeping the total number of neurons low. This strategy would be particularly advantageous for tasks characterized by small observation and action spaces, but would also imply a wider space of reservoirs HPs search in return.

\end{document}